\documentclass[runningheads]{llncs}

\usepackage{graphicx}
\usepackage{booktabs}
\usepackage{multirow}

\usepackage{placeins}
\usepackage[table]{xcolor}
\usepackage{colortbl}
\usepackage{booktabs}

\usepackage{subcaption}

\usepackage{algorithm}
\usepackage{algpseudocode}
\usepackage{array}

\usepackage{amsmath,amssymb}
\usepackage{xcolor}
\usepackage{hyperref}
\usepackage{caption}
\usepackage{tabularx}
\newcolumntype{Y}{>{\raggedright\arraybackslash}X}
\begin{document}

\title{ExpertoRhythm: Morphology-Aware Learning for Waveform Reconstruction and Cuffless Blood Pressure Estimation from Single-Channel PPG}
\titlerunning{Morphology-Aware Learning for Cuffless BP Estimation}


\author{Amir Arjomand \and
Kenneth B.~Kent \and
Georgiy Krylov}
\authorrunning{A. Arjomand et al.}

\institute{University of New Brunswick, Fredericton, Canada\\
\email{\{a.arjomand,ken,georgiy.krylov\}@unb.ca}}

\maketitle


\begin{abstract}
Continuous cuffless blood pressure (BP) monitoring from photoplethysmography (PPG) has strong potential for wearable health and telemonitoring, but accurate estimation remains difficult because PPG-to-BP mapping must preserve subtle waveform morphology and pressure-range-dependent dynamics. We introduce \textit{ExpertoRhythm}, an attention-enhanced 1D U-Net that reconstructs the arterial blood pressure (ABP) waveform from a single-channel PPG signal and derives systolic and diastolic BP directly from the reconstructed waveform. The central contribution is a composite morphology-aware learning objective that integrates range-weighted SmoothL1 reconstruction with a window-range regularizer to emphasize high-dynamic BP segments and reduce amplitude under/over-shoot. On the UCI cuff-less BP dataset with 942 subjects, ExpertoRhythm achieves 2.46/1.46~mmHg MAE for systolic/diastolic BP (SBP/DBP), while obtaining a 30.4\% average relative error reduction over pure MSE across waveform reconstruction and BP estimation metrics. Clinical-style evaluation further demonstrates low bias and strong agreement across the BP range, including high-pressure windows up to 200~mmHg, satisfying AAMI criteria and achieving BHS Grade~A. These results suggest that morphology-aware waveform reconstruction from a single PPG channel can provide an accurate and practical pathway toward continuous cuffless BP monitoring in wearable and remote-care settings.

\end{abstract}

\section{Introduction}
Blood pressure (BP) is a key marker of cardiovascular risk, but cuff-based measurements are inconvenient to take frequently and do not provide continuous trends. This has accelerated interest in \emph{cuffless} BP estimation using wearable-friendly sensors, especially photoplethysmography (PPG), which is inexpensive and already embedded in many consumer devices. A widely used deep-learning strategy is to reconstruct the arterial blood pressure (ABP) waveform from PPG and then derive systolic/diastolic values from the reconstructed waveform, retaining more hemodynamic detail than direct scalar regression \cite{ibtehaz2020ppg2abp,cheng2021}.

However, a practical mismatch remains between how these models are trained and how BP is ultimately read out. Most PPG$\rightarrow$ABP methods optimize pointwise waveform losses (MSE/MAE/SmoothL1), which improve average sample fidelity, but systolic/diastolic values are governed by localized morphology, here defined as peak/trough amplitudes and the immediate shape around those turning points. Consequently, small errors in these regions can yield noticeable BP error even when overall waveform error is low, making our loss a better surrogate by emphasizing these morphology critical points.

We address this gap with ExpertoRhythm, a channel-prior convolutional attention (CPCA) enhanced 1D U-Net for PPG$\rightarrow$ABP reconstruction \cite{huang2024channel,arjomand2025unetcpca} and composite training loss functions. ExpertoRhythm is trained with a composite loss that (i) range-weights a robust SmoothL1 loss to emphasize high-dynamic windows and (ii) applies a window-range regularizer to penalize amplitude under/over-shoot. On the UCI cuff-less BP dataset (942 subjects; 10\,s at 125\,Hz), controlled ablations show consistent improvements over standard pointwise objectives, achieving SBP/DBP MAE of 2.46/1.46\,mmHg while including high-pressure windows up to SBP $200$\,mmHg and meeting established medical evaluation standards (AAMI and BHS).

\paragraph{Contributions.}
\begin{itemize}
    \item We propose a morphology-aware training objective for PPG$\rightarrow$ABP reconstruction that combines range-weighted SmoothL1 with a window-range amplitude regularizer to improve waveform-derived SBP/DBP.
    \item We provide controlled ablations over both objective terms and model capacity, quantifying their effects on waveform fidelity and BP accuracy.
    \item We present a 4-layers CPCA 1D U-Net model and report standards-style evaluation demonstrating strong agreement under AAMI/BHS criteria.
\end{itemize}

\section{Background and Related Work}
\paragraph{Baselines and fair-comparison protocol.}
We benchmark our morphology-aware enhanced-CPCA 1D U-Net (ExpertoRhythm) against a focused, representative set of state-of-the-art PPG-to-ABP reconstruction methods that cover the dominant paradigms in the literature: convolutional encoder--decoder architectures (PPG2ABP~\cite{ibtehaz2020ppg2abp}, ABP-Net~\cite{cheng2021}), transformer-based models (our prior TransfoRhythm~\cite{arjomand2024transforhythm}), and recent morphology-aware or hybrid-objective approaches (De Palma~\emph{et al.}~\cite{de2025enhancing}).

These baselines were chosen because they (i) operate exclusively on single-channel PPG, (ii) target continuous waveform reconstruction for clinical SBP/DBP extraction, and (iii) are the most widely cited and directly comparable works under resource-constrained deployment scenarios.

For fair and reproducible comparison, we use the performance numbers **as originally reported** in each paper and adopt the identical evaluation protocol (MAE on the waveform, absolute SBP/DBP errors, etc.) and test-set definitions described by the authors. Where public test splits or preprocessing details are available, we follow them exactly. This approach eliminates confounding factors from data partitioning or metric variations and isolates performance gains to architectural and objective choices.

\section{Method}

\subsection{Regression Architecture and I/O}
\label{sec:arch}
\begin{figure*}[t]
  \centering
  \includegraphics[width=1\textwidth]{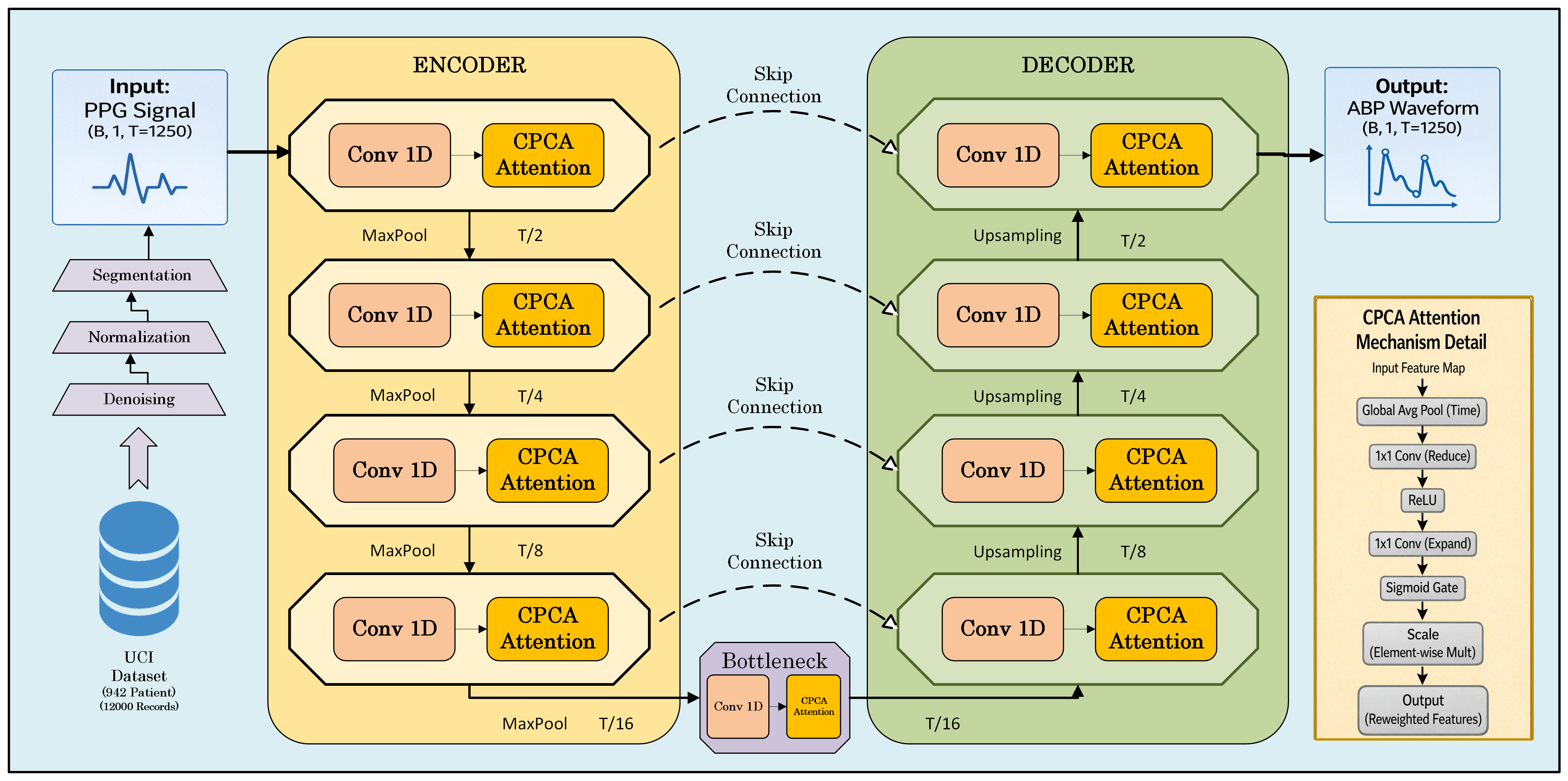}
  \caption{CPCA-enhanced 1D U-Net for PPG$\rightarrow$ABP regression (4-level encoder--decoder with skip connections and CPCA blocks).}
  \label{fig:architecture}
\end{figure*}

We reconstruct ABP from PPG using a 1D U-Net encoder--decoder with skip connections for time-series translation \cite{ronneberger2015u,stoller2018wave}. The network follows a 4-level multi-scale design as illustrated in Fig.~\ref{fig:architecture}.Concretely, with base width $C{=}72$, encoder widths are $\{C,2C,4C,8C\}$ with bottleneck $16C$, and the decoder mirrors this with skip concatenation.


\begin{itemize}
    \item \textbf{Encoder:} progressively downsamples temporal resolution using $\mathrm{MaxPool1D}(2)$ to learn hierarchical representations.
    \item \textbf{Bottleneck:} operates at the lowest resolution, aggregating long-range context across the full window into a compact representation that guides the decoder.
    \item \textbf{Decoder:} upsamples using nearest-neighbor interpolation and concatenates corresponding encoder features (skip connections) to recover fine-grained detail.
\end{itemize}

Each encoder and decoder stage uses the same lightweight block:
\begin{itemize}
    \item \textbf{Convolutional block:} two $\mathrm{Conv1D}$ layers (kernel size $3$) with ReLU and batch normalization \cite{kiranyaz20211d}.
    \item \textbf{CPCA attention:} appended at the end of each block to adaptively reweight channels using a squeeze-and-excitation style mechanism \cite{hu2018squeeze,si2025scsa,arjomand2025unetcpca}.
\end{itemize}

All convolutions use same-length padding, preserving temporal resolution within each scale and ensuring alignment for skip concatenation. Model capacity is controlled by the base number of channels in the first stage; we evaluate multiple widths in an ablation study in the Results section.Inputs are provided in channel-first form as $X\in\mathbb{R}^{B\times 1\times T}$, and the network outputs a time-aligned ABP estimate $\hat{Y}\in\mathbb{R}^{B\times 1\times T}$ for each window.Here, $B$ is the batch size, $T$ is the window length (samples), and $t$ indexes time.
For window-level reporting, SBP/DBP are computed from the predicted waveform extrema, consistent with prior PPG$\rightarrow$ABP reconstruction work \cite{ibtehaz2020ppg2abp,Athaya2021UNetABP,arjomand2025unetcpca}:
\begin{equation}
\widehat{\mathrm{SBP}}=\max_{t\in[1,T]}\hat{y}_t,\qquad
\widehat{\mathrm{DBP}}=\min_{t\in[1,T]}\hat{y}_t.
\end{equation}

\section{Morphology-Aware Loss}
\label{sec:loss}

Pointwise MAE/MSE can under-emphasize high-dynamic windows, which are critical for preserving ABP amplitude and extrema-derived SBP/DBP \cite{ibtehaz2020ppg2abp,Athaya2021UNetABP}. We combine a range-weighted SmoothL1 base loss with a window-range regularizer.

\paragraph{Range-weighted SmoothL1.}
For window $b$, let $r_b=\max(Y_b)-\min(Y_b)$ and $\bar r=\frac{1}{B}\sum_{j=1}^B r_j$.
We compute a detached importance weight
\begin{equation}
w_b=\frac{r_b}{\bar r+\epsilon},\qquad w_b\leftarrow \min(w_b,w_{\max}),
\end{equation}
and minimize
\begin{equation}
\mathcal{L}_{\text{base}}=\frac{1}{B}\sum_{b=1}^B w_b\,\mathrm{SmoothL1}_{\beta}(\hat{Y}_b,Y_b),
\end{equation}
where $\mathrm{SmoothL1}_{\beta}(\hat{Y}_b,Y_b)$ \cite{Huber_1964} denotes the mean over time samples in the window. Here $\epsilon$ ensures numerical stability (e.g., $r_b\!\approx\!0$), and $w_{\max}\in[3,5]$ prevents rare high-range windows from dominating updates.

\paragraph{Units.}
Losses are computed on min--max normalized ABP, while metrics are reported in mmHg after inverse scaling; thus $\delta$ is in normalized units (mmHg equivalent: $\delta\,(y_{\max}-y_{\min})$).

\paragraph{Range regularizer.}
To preserve within-window amplitude, after trimming $\tau$ samples at both ends we define
$\Delta R_b = (\max\hat{y}_b-\min\hat{y}_b) - (\max y_b-\min y_b)$
(over the trimmed interval) and penalize under/over-shoot beyond margin $\delta$ via a smooth gate:
\begin{equation}
\mathcal{L}_{\text{AR}}=\frac{1}{B}\sum_{b=1}^B\Big(
\sigma(s(-\Delta R_b-\delta))\,\mathrm{ReLU}(-\Delta R_b)^2+
\sigma(s(\Delta R_b-\delta))\,\mathrm{ReLU}(\Delta R_b)^2
\Big),
\end{equation}
where $\sigma(z)=1/(1+e^{-z})$ and $s>0$ controls gate steepness.

\paragraph{Overall.}
\begin{equation}
\mathcal{L}=\mathcal{L}_{\text{base}}+\lambda_{\text{AR}}\mathcal{L}_{\text{AR}}.
\end{equation}

\paragraph{Hyperparameters.}
Unless stated otherwise, we set SmoothL1 $\beta=1.0$, $\epsilon=10^{-6}$, $w_{\max}=5$, and trim $\tau=10$ samples per side; the range regularizer uses $\delta=0.01$ and gate slope $s=10$.\cite{PyTorchSmoothL1Loss}

\section{Data and Training Protocol}
\label{sec:data_train}
We use the widely used UCI dataset, derived from the MIMIC Waveform Database \cite{kachuee2015cuff,goldberger2000,saeed2011mimic}. It contains synchronized PPG and invasive ABP sampled at 125\,Hz (12{,}000 records from 942 subjects), segmented into 10\,s windows ($T{=}1250$).

Records are stored contiguously per subject; to prevent leakage, we split at the subject level by treating each subject’s block as indivisible. We keep a fixed 25\% held-out test split(random seed) and run 5-fold cross-validation on the remaining 75\% for model selection(per fold: 60\% train, 15\% validation, 25\% test); after choosing hyperparameters, we retrain on the full 75\% and evaluate once on the test set.

Preprocessing includes edge trimming and a 4th-order Butterworth bandpass (0.1--30\,Hz) with zero-phase filtering. We apply global min--max normalization separately for PPG and ABP; extrema are computed on each fold’s training split and applied to its validation data, and for the final model computed on the full 75\% pool and reused for the test set.

Models are implemented in PyTorch and trained on an RTX 3060 using batch size $B{=}256$ and Adam (lr $10^{-3}$, weight decay $10^{-5}$) for up to 150 epochs with early stopping on validation SBP/DBP MAE (patience 10) and ReduceLROnPlateau (factor 0.5). We set $\lambda_{\text{AR}}{=}0.6$, $w_{\max}{=}5$, and $\epsilon{=}10^{-6}$.

\section{Results}
\label{sec:results}

\subsection{Ablation Study}

We ablate two axes: model capacity (base width $C$) and objective design (morphology-aware loss). We first sweep $C$ under a fixed protocol, then fix the chosen width for loss ablations to isolate objective effects. All runs share identical splits, preprocessing, optimization, and metrics.


\paragraph{Architecture ablation (Table~\ref{tab:arch_ablation_channels}).}
We vary only the base channel width $C$. Increasing capacity from $C{=}32$ (prior work~\cite{arjomand2025unetcpca}) to $C{=}72$ clearly reduces SBP/DBP MAE, while $C{=}86$ yields only marginal gains at much higher cost. We therefore use $C{=}72$ for the best accuracy and size trade-off.

\begin{table}[t]
\centering
\footnotesize
\scriptsize
\setlength{\tabcolsep}{4pt}
\renewcommand{\arraystretch}{1.05}
\caption{Channel-width with MSE loss ablation}
\label{tab:arch_ablation_channels}
\begin{tabular}{c c c c}
\hline
Base channels ($C$) & Params (M) & Model size (MB) & SBP/DBP MAE (mmHg) \\
\hline
32 & 2.58 & 9.54  & 4.92 / 2.97 \\
48 & 5.63 & 21.46 & 4.41 / 3.09 \\
72 & 12.66 & 48.31 & \textbf{3.86 / 2.38} \\
86 & 18.06 & \textbf{68.89} & 3.78 / 2.26 \\
\hline
\end{tabular}
\end{table}

\paragraph{Loss ablation (Table~\ref{tab:loss_ablation}).}
On fixed architecture ($C=72$), range-weighted SmoothL1 ($\mathcal{L}_{\text{base}}$) outperforms MSE on waveform and BP errors. Adding $\mathcal{L}_{\text{AR}}$ gives further gains, peaking at $\lambda_{\text{AR}}=0.6$; higher values cause slight over-regularization.



\begin{table}[t]
\centering
\caption{Loss ablation. $\mathcal{L}_{\text{base}}$: range-weighted SmoothL1; $\mathcal{L}_{\text{AR}}$: range regularizer. Wave: MAE/RMSE; SBP/DBP from ABP extrema (mmHg). Improvement is the average relative error reduction versus pure MSE across Wave MAE, Wave RMSE, SBP MAE, and DBP MAE.}
\label{tab:loss_ablation}
\footnotesize
\scriptsize
\setlength{\tabcolsep}{3.5pt}
\renewcommand{\arraystretch}{1.05}
\begin{tabular}{l c c c c c}
\toprule
Loss setting & $\lambda_{\text{AR}}$ & Wave (MAE/RMSE) & SBP MAE & DBP MAE & Improvement vs. MSE \\
\midrule
MSE (\texttt{MSELoss}) & 0.0 & 4.68 / 6.97 & 3.86 & 2.38 & -- \\
$\mathcal{L}_{\text{base}}$ & 0.0 & 4.37 / 6.69 & 3.51 & 2.19 & +6.9\% \\
$\mathcal{L}_{\text{base}} + \lambda_{\text{AR}}\mathcal{L}_{\text{AR}}$ & 0.2 & 4.09 / 6.51 & 3.17 & 1.95 & +13.8\% \\
$\mathcal{L}_{\text{base}} + \lambda_{\text{AR}}\mathcal{L}_{\text{AR}}$ & 0.4 & 3.67 / 6.27 & 2.78 & 1.70 & +22.0\% \\

\arrayrulecolor{red}\specialrule{1pt}{0pt}{0pt}\arrayrulecolor{black}
\multicolumn{1}{!{\color{red}\vrule width 1pt}l}{
$\mathcal{L}_{\text{base}} + \lambda_{\text{AR}}\mathcal{L}_{\text{AR}}$}
& \textbf{0.6}
& \textbf{3.13 / 6.02}
& \textbf{2.46}
& \textbf{1.46}
& \multicolumn{1}{c!{\color{red}\vrule width 1pt}}{\textbf{+30.4\%}} \\
\arrayrulecolor{red}\specialrule{1pt}{0pt}{0pt}\arrayrulecolor{black}

$\mathcal{L}_{\text{base}} + \lambda_{\text{AR}}\mathcal{L}_{\text{AR}}$ & 0.8 & 3.16 / 6.05 & 2.52 & 1.48 & +29.6\% \\
\bottomrule
\end{tabular}
\end{table}

\subsection{Metrics and Clinical Result}

\begin{figure}[t]
    \centering
    \begin{subfigure}[t]{0.49\linewidth}
        \centering
        \includegraphics[width=0.6\linewidth]{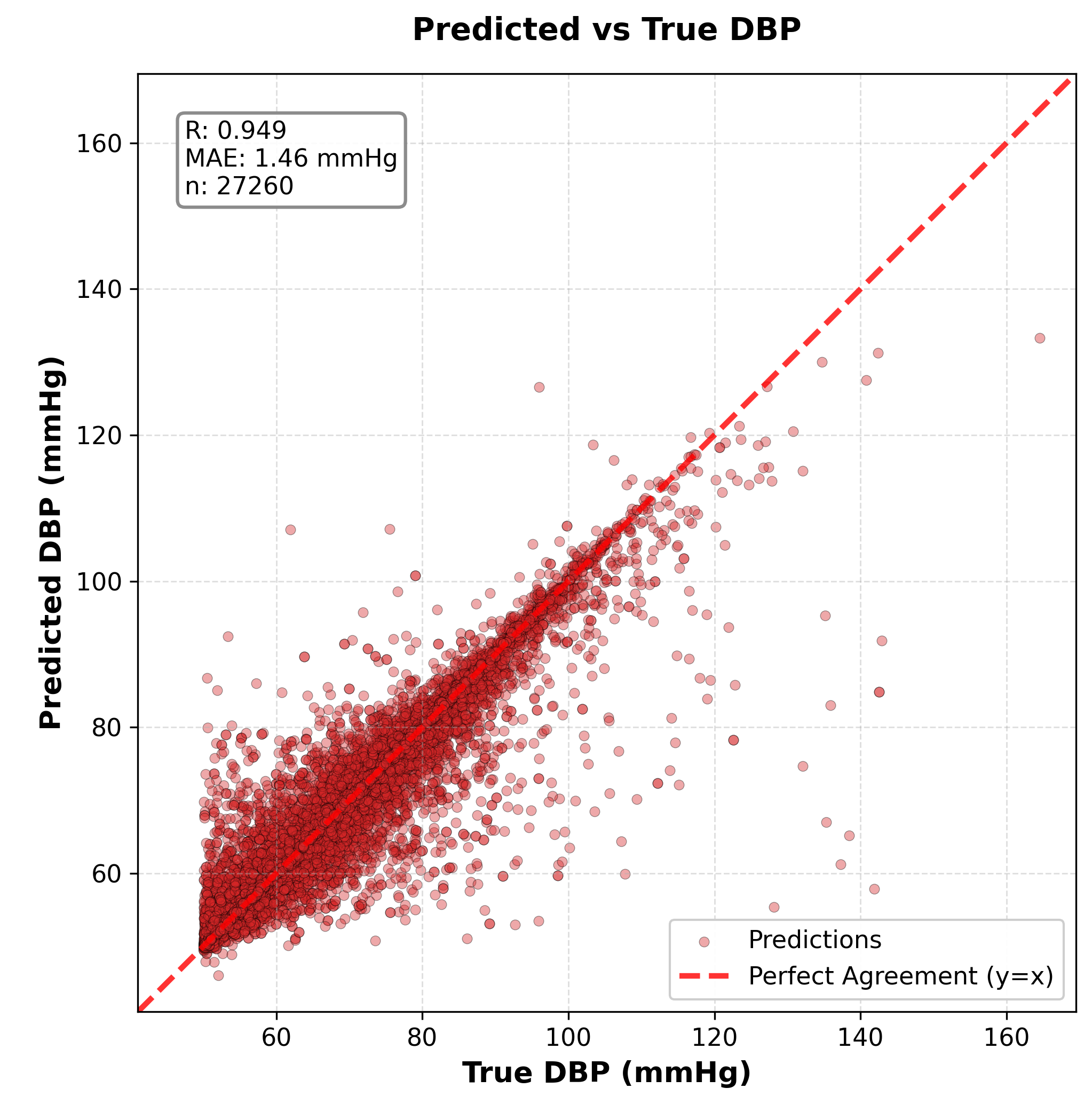}
        \caption{DBP}
        \label{fig:sctr_dbp}
    \end{subfigure}\hspace{1mm}
    \begin{subfigure}[t]{0.49\linewidth}
        \centering
        \includegraphics[width=0.6\linewidth]{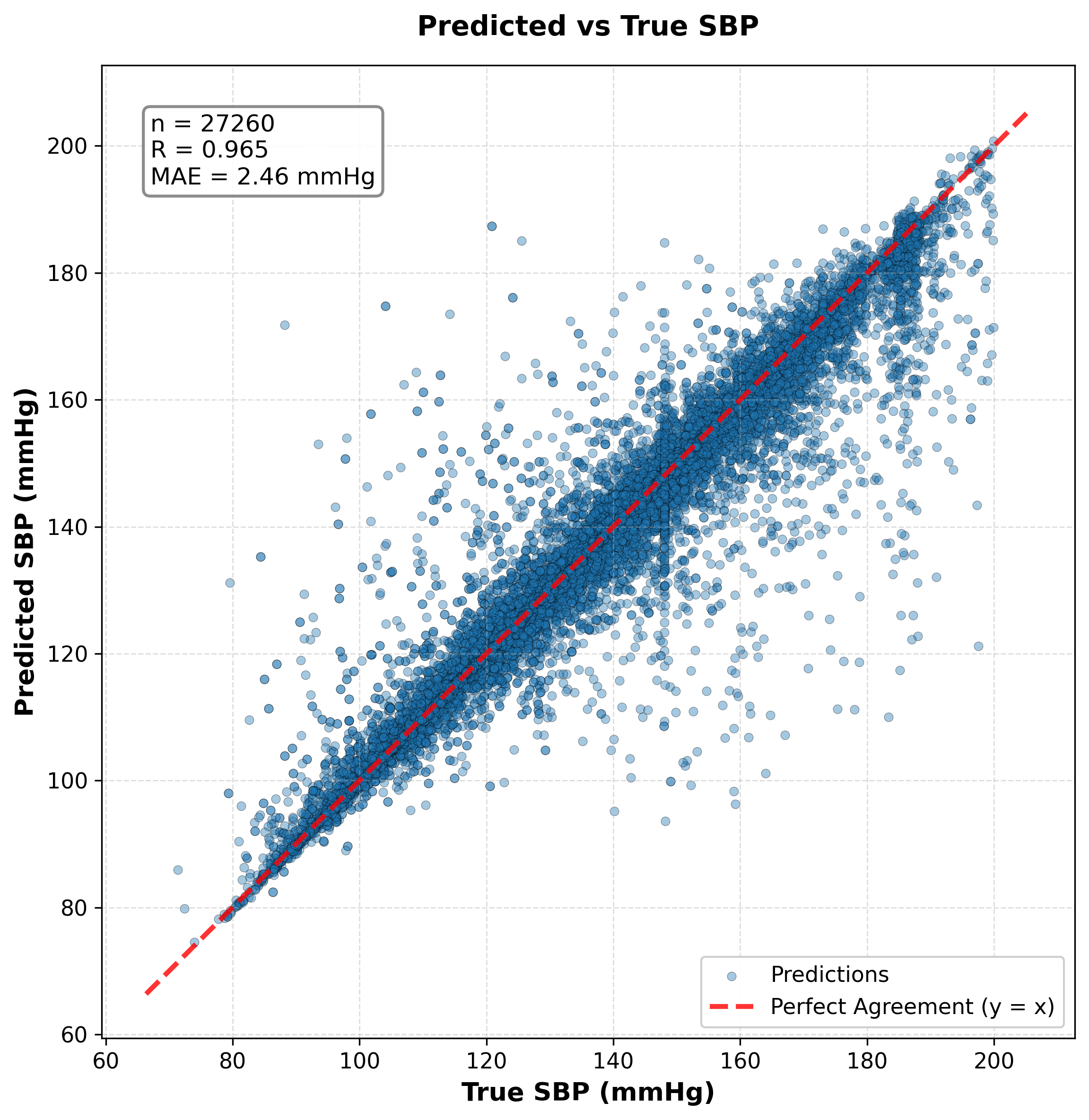}
        \caption{SBP}
        \label{fig:sctr_sbp}
    \end{subfigure}
    \caption{Predicted versus reference blood pressure on the test set. The dashed line indicates the identity line.}
    \label{fig:sctr_bp}
\end{figure}

\begin{figure}[t]
    \centering
    \includegraphics[width=0.70\linewidth,trim=10 8 10 8,clip]{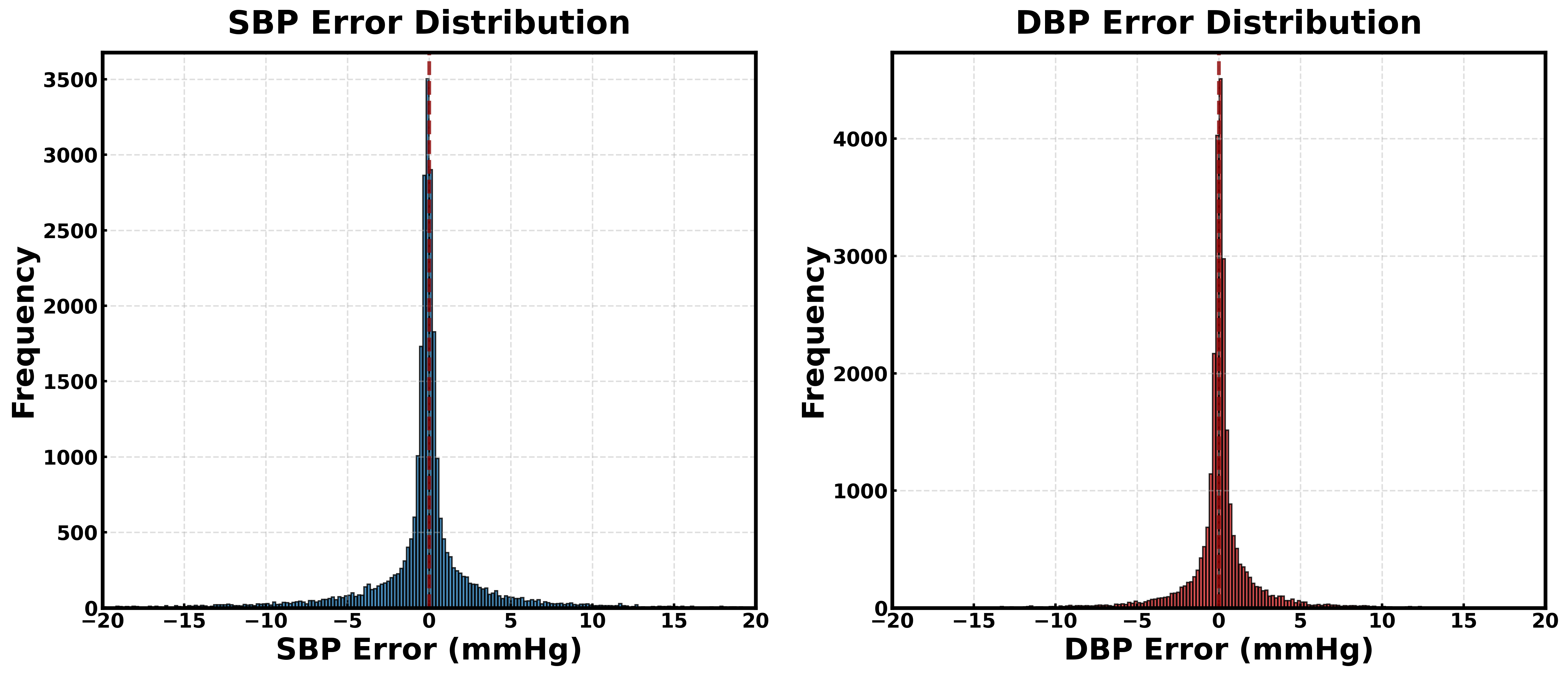}
    \caption{\vspace{-1mm}Error distributions for SBP and DBP on the test set.\vspace{-2mm}}
    \label{fig:error_hist}
\end{figure}

\begin{figure}[t]
    \centering
    \includegraphics[width=0.70\linewidth]{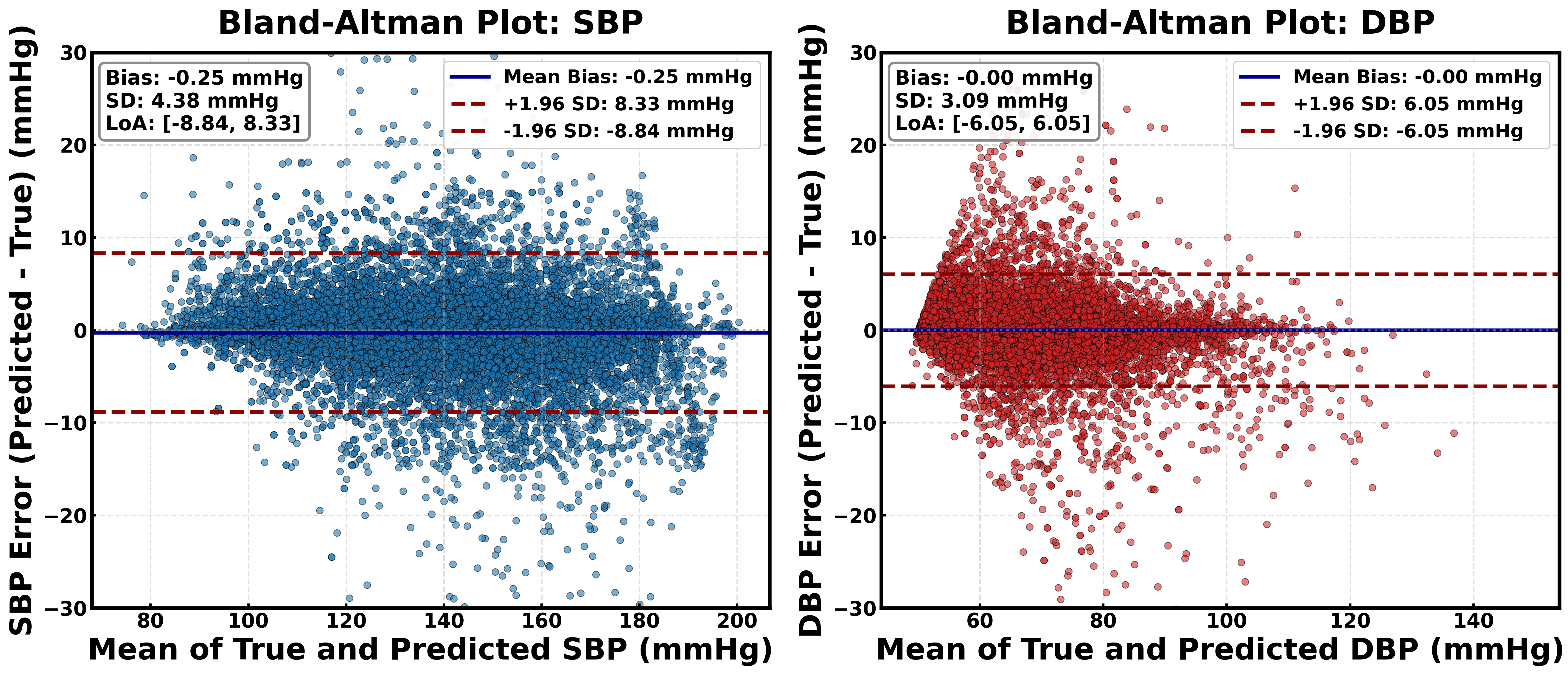}
    \caption{Bland--Altman plots for SBP/DBP. Solid: mean bias; dashed: 95\% LoA (mean $\pm 1.96\,\mathrm{SD}$).}
    \label{fig:bland_altman}
\end{figure}

Across the held-out test set, the proposed model achieves low waveform reconstruction error and strong clinical agreement when BP values are extracted from predicted ABP extrema. The errors derived from extrema are small and near-unbiased: SBP MAE is 2.46~mmHg (mean error $-0.37$~mmHg, SD 4.38~mmHg) and DBP MAE is 1.46~mmHg (mean error $-0.07$~mmHg, SD 3.63~mmHg), with error histograms tightly centered around zero (Fig.~\ref{fig:error_hist}). Predicted-versus-true scatter plots further confirm strong agreement with the identity line, with high correlation for both DBP ($R=0.949$, MAE 1.46~mmHg) and SBP ($R=0.965$, MAE 2.46~mmHg) over $n=27{,}260$ windows (Fig.~\ref{fig:sctr_dbp}, Fig.~\ref{fig:sctr_sbp}). Bland--Altman analysis shows near-zero bias and tight limits of agreement for both SBP and DBP, with no evident proportional bias across the pressure range (Fig.~\ref{fig:bland_altman})
. Finally, using established medical standards (AAMI and BHS), our evaluation (Table~\ref{tab:clinical_eval}) meets AAMI mean/SD requirements for both SBP and DBP and achieves BHS Grade~A for both pressures. Errors are near-unbiased with variability within AAMI limits, and the BHS threshold criteria are satisfied across the standard tolerance bands.

\subsection{Comparison to State-of-the-Art}
\label{sec:comparison}

Table~\ref{tab:bp_models_sorted_dataset} compares representative cuffless BP methods while highlighting key comparability factors. ExpertoRhythm achieves $2.46/1.46$\,mmHg SBP/DBP MAE using a single raw PPG channel, improving over widely used UCI baselines such as Dual-DS-U-Net~\cite{ibtehaz2020ppg2abp} and ResNet34~\cite{qin2023cuff} and remaining competitive with recent sequence models~\cite{fan2025dual}. Several strong results rely on richer representations (e.g., PPG+STFT~\cite{tian2026}) or additional derived inputs (PPG/VPG/APG~\cite{pan2024}), whereas our gains are achieved under a minimal sensing constraint by optimizing for peak-sensitive morphology via a composite loss. Notably, U-Net+BiLSTM~\cite{pan2024} reports SBP/DBP errors close to ours, but it is evaluated on a substantially smaller subset (150 records versus 12{,}000), which can materially affect difficulty and comparability. The MIMIC-III entries also suggest that accuracy depends strongly on modality and preprocessing (e.g., ECG+PPG or engineered features~\cite{shen2024}), motivating PPG-only methods that better preserve waveform structure.

\begin{table*}[t]
\centering
\caption{Comparison of cuffless BP estimation models. ND = not disclosed.}
\label{tab:bp_models_sorted_dataset}
\tiny
\setlength{\tabcolsep}{2pt}
\renewcommand{\arraystretch}{0.95}
\begin{tabularx}{\textwidth}{@{}X l r X c c r c@{}}
\toprule
Model & Feature & \#Patients & Input & SBP & DBP & Records & Comp. Loss \\
\midrule
\multicolumn{8}{@{}l}{\textbf{MIMIC-III}}\\
ABP-Net~\cite{cheng2021} & Raw & 106 & PPG/VPG/APG & 3.27 & 1.90 & ND & Yes \\
ArterialNet~\cite{huang2025arterialnet} & Raw & 61 & PPG & 4.18 & 5.17 & ND & Yes \\
Ensemble\_NN~\cite{shen2024} & FE & 96 & ECG+PPG & 3.74 & 2.22 & ND & Yes \\
MLR~\cite{lin2021towards} & FE & 109 & PPG/VPG/APG & 4.59 & 2.47 & 12000 & No \\
Residual U-Net~\cite{de2025enhancing} & Raw & 120 & PPG & 13.12 & 5.48 & ND & Yes \\
\midrule
\multicolumn{8}{@{}l}{\textbf{UCI(MIMIC-II extension)}}\\
Dual-DS-U-Net~\cite{ibtehaz2020ppg2abp} & Raw & 942 & PPG & 5.73 & 3.45 & 12000 & No \\
Transformer~\cite{fan2025dual} & Raw & 942 & PPG & 2.97 & 1.60 & 12000 & No \\
LSTM autoenc.~\cite{harfiya2021continuous} & Raw & 942 & PPG & 4.05 & 2.41 & 5289 & No \\
MCAFNet~\cite{tian2026} & Freq & 942 & PPG+STFT & 2.48 & 1.29 & ND & No \\
ResNet34~\cite{qin2023cuff} & Raw & 942 & PPG & 5.98 & 3.24 & 12000 & No \\
U-Net+BiLSTM~\cite{pan2024} & Raw & ND & PPG/VPG/APG & 2.48 & 1.42 & 150 & Yes \\
\midrule
ExpertoRhythm (ours) & Raw & 942 & PPG & 2.46 & 1.46 & 12000 & Yes \\
\bottomrule
\end{tabularx}
\end{table*}

\begin{table}[!t]
\centering
\caption{Clinical evaluation under AAMI and BHS.}
\label{tab:clinical_eval}
\tiny
\setlength{\tabcolsep}{3pt}
\renewcommand{\arraystretch}{0.92}
\begin{tabular}{l c c c c c c}
\toprule
\multirow{2}{*}{Metric} &
\multicolumn{3}{c}{SBP} &
\multicolumn{3}{c}{DBP} \\
\cmidrule(lr){2-4}\cmidrule(lr){5-7}
 & Val. & Crit. & Res. & Val. & Crit. & Res. \\
\midrule
\multicolumn{7}{@{}l}{\textbf{AAMI}}\\
ME (mmHg) & -0.36 & $\le 5$ & PASS & -0.07 & $\le 5$ & PASS \\
SD (mmHg) &  4.38 & $\le 8$ & PASS &  3.63 & $\le 8$ & PASS \\
AAMI overall & \multicolumn{3}{c}{\textbf{PASS}} & \multicolumn{3}{c}{\textbf{PASS}} \\
\midrule
\multicolumn{7}{@{}l}{\textbf{BHS}}\\
$\le\pm5$ (\%)  & 87.2 & $\ge 60$ & A & 93.3 & $\ge 60$ & A \\
$\le\pm10$ (\%) & 94.6 & $\ge 85$ & A & 97.6 & $\ge 85$ & A \\
$\le\pm15$ (\%) & 96.9 & $\ge 95$ & A & 98.9 & $\ge 95$ & A \\
BHS grade & \multicolumn{3}{c}{\textbf{A}} & \multicolumn{3}{c}{\textbf{A}} \\
\bottomrule
\end{tabular}
\end{table}


\section{Discussion and Conclusion}
The results highlight two practical drivers of performance: model capacity and objective alignment. The channel-width sweep (Table~\ref{tab:arch_ablation_channels}) shows that increasing width improves SBP/DBP accuracy up to a moderate setting, after which gains saturate. Holding architecture fixed, the loss ablation (Table~\ref{tab:loss_ablation}) demonstrates that the proposed morphology-aware objective consistently reduces SBP/DBP error by improving amplitude precise waveform reconstruction, rather than relying on architectural scaling alone. Further analyses support clinical credibility under established medical evaluation standards (Table~\ref{tab:clinical_eval}). Overall, ExpertoRhythm shows that the waveform can be recovered from a single PPG channel when the training objective is designed to preserve it, achieving strong accuracy and meeting clinical grade criteria. A remaining limitation is deployability: wearable use requires low-latency, energy-efficient edge inference. Future work will report efficiency metrics (latency, memory, compute/energy proxies) and study deployment-oriented optimization, including quantization and hybrid mixtures-of-experts for dynamic compute. Also we will validate ExpertoRhythm on MIMIC-III, MIMIC-IV, and wearable-acquired PPG datasets to assess cross-dataset generalizability using consistent subject-level protocols.

\FloatBarrier

\section*{Acknowledgements} Funding for this research was provided by ResearchNB through the Academic Start-up Fund, formerly the Talent Recruitment Fund program (TRF 2025 005). The authors gratefully acknowledge this support. This preprint has not undergone any post-submission improvements or corrections. The Version of Record of this contribution is published in Volume 16748 of the Lecture Notes in Artificial Intelligence (LNAI) series, a subseries of Lecture Notes in Computer Science (LNCS), and is available online at \url{https://doi.org/10.1007/978-3-032-30710-1_39}.

\subsubsection*{Data and Code Availability}
The source code used in this study is available at: \url{https://github.com/amirarjmand93/aime2026_loss_function}.

\bibliographystyle{splncs04}
\bibliography{references}

\end{document}